\documentclass[letterpaper, 10 pt, journal, twoside]{IEEEtran}  % Comment this line out if you need a4paper

\IEEEoverridecommandlockouts                              % This command is only needed if 
\usepackage{graphicx} % for pdf, bitmapped graphics files
\usepackage{amsmath} % assumes amsmath package installed
\usepackage{amssymb}  % assumes amsmath package installed
\usepackage[usenames, dvipsnames]{color}
\usepackage{siunitx}
\usepackage{cases}
\usepackage{lipsum}
\usepackage{color}
\usepackage{cite}
\usepackage[ruled,vlined]{algorithm2e}
\usepackage{diagbox} 
\usepackage{tabu}
\usepackage{array}
\usepackage{multirow}
\usepackage{booktabs}
\usepackage{makecell}
\usepackage{empheq}
\usepackage{threeparttable}
\usepackage{tabularx}

\newcolumntype{P}[1]{>{\centering\arraybackslash}p{#1}}
\newcolumntype{M}[1]{>{\centering\arraybackslash}m{#1}}

\DeclareMathOperator*{\argmax}{arg\,max}

\newcommand{\revision}[1]{\textcolor{black}{#1}} %for displaying red texts
\newcolumntype{C}[1]{>{\centering\let\newline\\\arraybackslash\hspace{0pt}}m{#1}}
\newcolumntype{L}[1]{>{\raggedright\let\newline\\\arraybackslash\hspace{0pt}}m{#1}}
\title{
Autonomous Path Planning for Intercostal Robotic Ultrasound Imaging Using Reinforcement Learning
}

\author{Yuan Bi*$^{1}$, Cheng Qian*$^{1}$, Zhicheng Zhang$^{2}$, Nassir Navab$^{1}$, \textit{Fellow, IEEE}, and Zhongliang Jiang$^{1}$ % <-this % stops a space

\thanks{$^{*}$ Authors are with equal contributions.}
\thanks{Corresponding author: Zhongliang Jiang.}
\thanks{$^{1}$Y. Bi, C. Qian, N. Navab, and Z. Jiang are with the Chair for Computer-Aided Medical Procedures and Augmented Reality, Technical University of Munich, Boltzmannstr. 3, 85748 Garching bei M\"unchen, Germany (zl.jiang@tum.de)     }%% 
\thanks{$^{2}$Z. Zhang is with the Software and Societal Systems Department, School of Computer Science, Carnegie Mellon University, 5000 Forbes Avenue, Pittsburgh, PA, United States}
}

\begin{document}

\maketitle

%%%%%%%%%%%%%%%%%%%%%%%%%%%%%%%%%%%%%%%%%%%%%%%%%%%%%%%%%%%%%%%%%%%%%%%%%%%%%%%%
\begin{abstract}
% Ultrasound (US) is frequently used in clinical practice for internal organ screening and intervention guidance thanks to its high portability. However, the flexibility of US also induces high inter- and intra-operator variations. Benefiting from the high reproducibility of robots, robotic US system (RUSS) appears to be a promising solution.
Ultrasound (US) has been widely used in daily clinical practice for screening internal organs and guiding interventions. However, due to the acoustic shadow cast by the subcutaneous rib cage, the US examination for thoracic application is still challenging. To fully cover and reconstruct the region of interest in US for diagnosis, an intercostal scanning path is necessary.
% Nonetheless, US imaging suffers from inter- and intra-operator variations. Leveraging the reproducibility offered by robots, robotic US systems emerge as a promising solution.
% However, scanning path planning for autonomous robotic US remains challenging, especially in intercostal regions with limited acoustic windows. 
To tackle this challenge, we present a reinforcement learning (RL) approach for planning scanning paths between ribs to monitor changes in lesions on internal organs, such as the liver and heart, which are covered by rib cages. Structured anatomical information of the human skeleton is crucial for planning these intercostal paths. To obtain such anatomical insight, a RL agent is trained in a virtual environment constructed using computational tomography (CT) templates with randomly initialized tumors of various shapes and locations.
% To do so, experienced sonographers are proficient at utilizing anatomical knowledge, such as the skeleton information, to simplify the process. 
% Illuminated by this insight, an RL agent is trained in a virtual environment built on computational tomography (CT) templates to generate scanning trajectory for full coverage of target volumes beneath the rib cage. 
% A simulation environment for intercostal US scans is built in Blender for both training and testing purposes. 
In addition, task-specific state representation and reward functions are introduced to ensure the convergence of the training process while minimizing the effects of acoustic attenuation and shadows during scanning. 
% The planned trajectory can then be registered to the patient on-site via established skeleton-based registration methods.
To validate the effectiveness of the proposed approach, experiments have been carried out on unseen CTs with randomly defined single or multiple scanning targets. The results demonstrate the efficiency of the proposed RL framework in planning non-shadowed US scanning trajectories in areas with limited acoustic access.
% The ablation study demonstrates the efficiency of the RL agent in planning non-shadowed trajectories in the area with limited acoustic access. 
% The performance of the proposed method when facing multiple scanning targets is also validated in the test session.
% Furthermore, we validate the performance of the proposed method when confronted with multiple scanning targets during the test phase.
\end{abstract}

% \markboth{IEEE Robotics and Automation Letters. Preprint Version. Accepted April, 2022}
% {Bi \MakeLowercase{\textit{et al.}}: VesNet-RL: Simulation-based Reinforcement Learning for Real-World US Probe Navigation} 

%%%%%%%%%%%%%%%%%%%%%%%%%%%%%%%%%%%%%%%%%%%%%%%%%%%%%%%%%%%%%%%%%%%%%%%%%%%%%%%%
% \vspace{1em}
% %%%%%%%%%%%%%%%%%%%%%%%%%%%%%%%%%%%%%%%%%%%%%%%%%%%%%%%%%%%%%%%%%%%%%%%%%%%%%%%%

%%%%%%%%%%%%%%%%%%%%%%%%%%%%%%%%%%%%%%%%%%%%%%%%%%%%%%%%%%%%%%%%%%%%%%%%%%%%%%%%
\bstctlcite{IEEEexample:BSTcontrol}
 \section{Introduction}\label{sec:intro}
% \lipsum[2-3]
%%%%%%%%%%%%%%%%%%%%%%%%%%%%%%%%%%%%%%%%%%%%%%%%%%%%%%%%%%%%%%%%%%%%%%%%%%%%%%%%%%%

%Ultrasound in general

\begin{figure}[t!]
\centering
\includegraphics[width=0.48\textwidth]{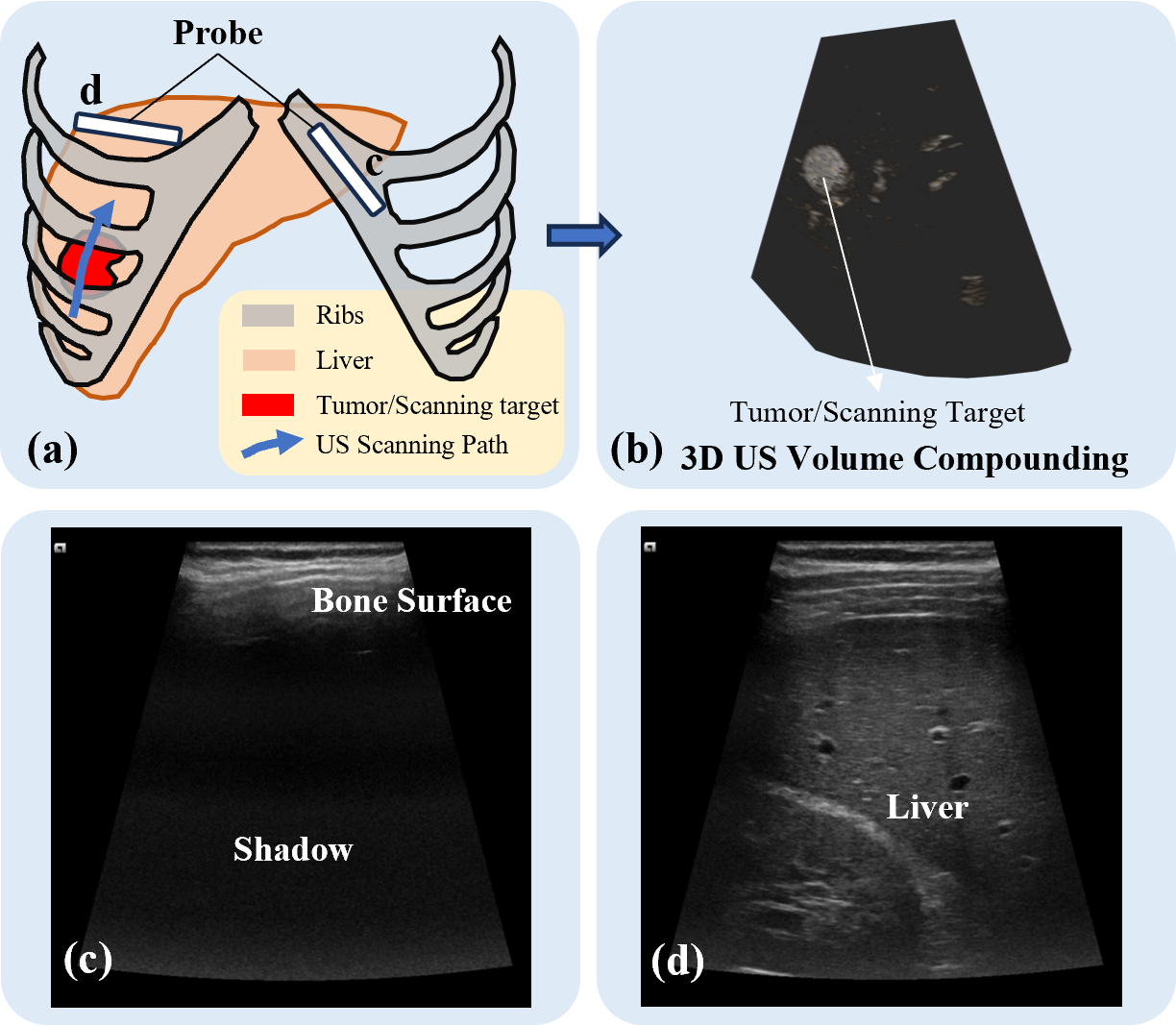}
\caption{(a) Our target is to plan a US scanning trajectory to fully cover a specific area under ribs. Such area can be an already identified tumor or some suspicious regions defined by the doctors based on CT images. (b) Combining with the tracking information, the final goal is to realize the US reconstruction of the selected area. The planned trajectory should avoid the occlusion of bones and try to perform the US acquisition through intercostal gaps. Two representing probe positions are shown in (a) and the resulting US images are displayed in (c) and (d), respectively.
}
\label{Fig_problem}
\end{figure}

% \IEEEPARstart{I}{N} 
Ultrasound (US) imaging has been one of the most widely used modalities in clinical practice, e.g., for fetuses~\cite{fiorentino2022deep}, internal organs~\cite{mustafa2013development}, and vessels~\cite{jiang2022towards}. It has shown superiority of being low-cost, high portable, and real-time in comparison to computed tomography (CT) and magnetic resonance imaging (MRI). 
% US  has been widely used for examining internal organs, e.g., liver, kidney, arteries, etc. 
Benefiting from its real-time characteristic, US is also frequently used to provide needle guidance for biopsy or ablation procedures (e.g., liver, prostate, breast, etc.). As a portable device, US gives operators high flexibility during examinations, but in the meantime, it also hinders the standardization of image acquisitions. A lot of factors can affect the quality of US images, including internal (frequency of US wave, dynamic range, focus, etc.) and external acquisition parameters (applied force and pose of the probe)~\cite{ipsen2021towards,si2024design,dyck2024towards}. In addition, to monitor the anatomy of interest covered by the rib cage (e.g., liver or heart), a proper scanning path in the limited intercostal space is needed to avoid acoustic shadows cast by bone structures [see Fig.~\ref{Fig_problem} (c)]. For instance, during and after the liver ablations, the surgeons need to monitor the tumor as a whole to correctly assess the surgical outcomes. This task requires specific training to understand both the US images and underlying anatomical knowledge. Therefore, inter- and intra-operator variations often arise.

% Meanwhile, because of the high acoustic impedance of bone structures, the sonographers need to plan the US scanning path intercostally to avoid shadows created by bones when screening organs covered by rib cage, e.g., heart and liver. As a result, US imaging often suffers from high inter-operator variance and low reproducibility.

% \begin{figure}[ht!]
% \centering
% \includegraphics[width=0.45\textwidth]{images/introduction.png}
% \caption{Illustration of three representative US images for thoracic application. The resulting US images of the three US probe poses displayed in (a) are shown in (b), (c), and (d) respectively. (b) shows the case when the acoustic wave is fully blocked by the rib, while the US view in (c) is only partially hindered. (d) depicts the one with a clear US view through the intercostal space. 
% }
% \label{Fig_intro}
% \end{figure}

\par
Robotic systems have demonstrated their super-human performance in terms of high precision, stability, and reproducibility. The studies in developing robotic US systems (RUSS) offer a promising solution to overcome the aforementioned operator-dependent variations~\cite{jiang2023robotic, li2021overview}.
Based on the US propagation principle, keeping the US probe orthogonal to the surface can maximise the capturing of reflected acoustic signals, thus resulting in better visualisation results (i.e., contrast)~\cite{jiang2020automatic,akbari2021robot}. However, this is not the case for thoracic applications, where the visibility of internal organs will be significantly hindered due to the acoustic shadow generated by bones with high acoustic impedance.
Several rule-based path planning method has been proposed based on external camera~\cite{huang2024robot,huang2021towards,bal2023curvature,tan2023autonomous} or pre-operative CT~\cite{gobl2017acoustic}. These methods can perform robustly in their respective applications. However, given the complexity of the intercostal area, automatically generating a US scanning trajectory to fully cover and reconstruct a specific volume under the ribs while avoiding acoustic shadow remains challenging.

\par
To address this challenge, reinforcement learning (RL) has emerged as a promising alternative solution for serial decision-making problems. RL achieves this by interacting with the environment while trying to maximize the task-related reward functions. Specifically for the application of RUSS, RL has been used to search for standard US views with meaningful diagnostic information~\cite{li2021image,bi2022vesnet,hase2020ultrasound,deng2021learning}. However, the objective of the aforementioned methods is to find standard planes rather than planing a scanning path to reconstruct and monitor the changes of lesions (e.g., liver tumors) covered by rib cage. In addition, these studies only use partial observation, such as US images, contact force, and action history, rather than the true state, i.e., the relative position between the probe and the scanned objects. 
% However, considering sonographers will mentally map their observations to the anatomical atlas in their mind, a full observation can significantly reduce the complexity of using RL exploring an unkown environment. 
However, the ability to leverage anatomical knowledge as prior information, and mapping realtime observations to the virtual model in clinicians’ minds can cast the partially observable problem into a fully observable problem, thus reducing the complexity of the task. This is especially useful for complex navigation tasks of liver with limited intercostal acoustic windows.

% has gained rising attention. By interacting with the environment, RL agent can learn a policy to execute complex tasks based on observations at every state. 
% RL-based methods usually modeled the US navigation problem as a Partially Observable Markov Decision Process (POMDP), where the true state (relative position of the probe to the scanning target) can only be estimated through observations (US images, contact force, action history, etc.)~\cite{li2021image,bi2022vesnet,hase2020ultrasound,deng2021learning}. It is designed in such a way to mimic the navigation process of sonographers, who are tainted to do US examinations based on the image feedback. However, the ability to leverage anatomical knowledge as prior information, and mapping real-time observations to the virtual model in clinicians' minds can cast the partially observable problem into a fully observable problem, thus reducing the complexity of the task. This is especially useful for complex navigation tasks of the liver with limited intercostal acoustic windows.

\par
% In this study,
% Our aim is to find a trajectory that can reconstruct the selected region under ribs using US.
To generate the intercostal scanning path for properly visualizing volumes of interest beneath ribs, this study presents an RL framework trained based on CT templates. To effectively characterize the state representation at each step, the 3D scenery views of the rib cage, scanning target (i.e., liver tumors here), and the US imaging plane are jointly utilized. The use of the 3D scenery view casts the problem into a fully observable Markov decision process (FOMDP), which can provide global hints. Therefore, in this way, FOMDP can enhance the robustness of the trained policy for finding the proper scanning trajectory. The main contributions of this article are listed as follows:
\begin{itemize}
  \item An RL-based method is proposed to tackle the challenging task of intercostal US scanning path planning. Instead of US image guided trajectory planning, this approach uses CT atlases to cast a partially observable decision-making problem into a fully observable one. Task-specific reward functions and state representations are proposed to ensure full coverage and reconstruction of single or multiple regions of interest while minimizing the acoustic attenuation and shadows.
  \item A simulation environment for US screening of internal organs or lesions through intercostal spaces has been developed based on the CT atlas. A cylindrical coordinate system is employed to efficiently model the 6D movements of US probes in the vicinity of the ribcage region and ensure the adaption of patients with varying body sizes. 
\end{itemize}
Finally, the effectiveness of the proposed framework has been validated on publicly available unseen patients' data~\cite{soler20103d}. In the next step, to further map the planned path from CT template to individual patients for autonomous robotic scanning, the non-rigid skeleton graph-based registration algorithm~\cite{jiang2023skeleton,jiang2023thoracic} can be integrated. However, it is worth noting that this study primarily focuses on robust and autonomous path generation. The code will be released upon acceptance.
% \footnote{Code: https://github.com/yuan-12138/Intercostal-US-RL}  
% \footnote{The video: https://www.youtube.com/***}.

% \par
% A comprehensive autonomous US scanning system for liver should consist of different components: 1) a path planning module, 2) a registration module, 3) an execution module, and 4) an adjusting module. The planned scanning path is registered to the patient on-site and the robot is employed to execute the scanning while ensuring appropriate contact and force interaction with the surface. Considering the dynamical movements of the liver caused by breathing, an adjusting module based on the acquired US images is incorporated to mitigate and compensate for such variations. By applying non-rigid registration approaches~\cite{jiang2023skeleton,jiang2023thoracic}, we can then plan the trajectory on a CT atlas and realize non-patient-specific path projection.
% One possible application scenario for such robotic system is liver ablations. The robot can then help the surgeons to monitor the target tumor as a whole during and after the ablation to assess the surgical outcomes.
% This work, as a fundamental component within the broader system, is primarily devoted to designing a robust path planning algorithms tailored to the comprehensive framework.

\section{Related Work}
\subsection{Path Planning for RUSS}
In order to guarantee the full coverage of the anatomy of interest, Wang~\emph{et al.} introduced a path planning pipeline for breast US scanning based on point clouds captured by an external RGB-D camera~\cite{wang2022full}. 
Yang~\emph{et al.} utilized a deep learning network to extract the spine area of patients from camera input to pre-plan the US sweep trajectory~\cite{yang2021automatic}. 
% In this regard, Bal~\emph{et al.} proposed an evaluation score to assess the point clouds reconstructed from RGB-D camera for better US trajectory generation~\cite{bal2023curvature}.
Besides external cameras, CT and MRI are also frequently utilized for scanning path planning.
Hennersperger~\emph{et al.} proposed a registration-based method to transfer the scanning path planed on pre-operative MRI/CT images to patients~\cite{hennersperger2016towards}. 
Jiang~\emph{et al.} implemented an MRI atlas-based non-rigid registration framework to address the articulated motions of human arms~\cite{jiang2022towards}. 
These methods work robustly in their cases, while their effectiveness is decayed for thoracic applications due to the occlusion of rib cage.
% The aforementioned trajectory planning approaches are relatively intuitive since scanning environment is comparatively clean without interfering of bones. 

\par
In order to image the whole liver, Mustafa~\emph{et al.} localized the region of interest by extracting umbilicus and mammary papillae from a web camera and then the probe was controlled to perform wobbler motion to obtain a wider view~\cite{mustafa2013development}. However, its performance in intercostal spaces is limited without explicitly considering the bone occlusion. To address this challenge, G\"obl~\emph{et al.} developed an automatic path planning approach based on patient-specific CT to generate shadow-free acquisition poses for a given target point using conventional rule-based constraints~\cite{gobl2017acoustic}. Al-Zogbi~\emph{et al.} introduced a RUSS to automatically generate the US acquisition pose for a chosen landmark point in lung and the robot is controlled to perform a wobbler movement at the acquisition point to collect diagnosis images~\cite{al2021autonomous}. To eliminate the need for pre-operative CT, Shida~\emph{et al.} introduced a visual servoing searching method to find the acoustic window for parasternal long-axis view of heart~\cite{shida2023automated}. 
Nonetheless, instead of acquiring one specific US frame, how to automatically generate a US scanning trajectory to fully cover and reconstruct a specific volume under ribs still requires a more intelligent and advanced algorithm. Reinforcement learning (RL), on the other hand, provides a promising alternative.

% RL
\subsection{Reinforcement Learning for RUSS}
% Through enormous amount of interactions with the environment, RL agent is able to learn a policy to execute complex tasks based on observations at each state. 
The great successes in gaming have demonstrated the superiority of RL in solving sequential decision processes~\cite{mnih2015human}.
Multiple efforts have been made in the field of US imaging analysis by using RL for landmark detection~\cite{yang2021searching, alansary2019evaluating}, segmentation~\cite{sahba2008application}, and video assessment~\cite{liu2020ultrasound}. In addition, initial attempts have also been conducted to adopt RL for RUSS.
Hase~\emph{et al.} implemented an RL agent to guide a US probe for standard plane localization of spine~\cite{hase2020ultrasound}. 
For the similar application scenario, Li~\emph{et al.} enlarged the action space by taking the orientation into account~\cite{li2021image}. 
% Li~\emph{et al.} implemented an RL agent to control the 6-D pose of a US probe for standard plane localization of spine~\cite{li2021image}.
Based on the observation from an external camera, Ning~\emph{et al.} developed an RL system to automatically locate the standard US plane~\cite{ning2021autonomic}.
To increase the generalization ability of the trained RL model, Bi~\emph{et al.} utilized the segmented masks of the anatomy of interest as inputs to bridge the gap between simulation and real environment~\cite{bi2022vesnet}. 
% Ning~\emph{et al.} introduced an RL-based posture adjustment module based solely on contact force to maintain a specific contact angle between US probe and scanning surface~\cite{ning2021force}. 
% Combining the force-based RL module with a US-image based directional guidance network, the autonomous vascular scanning was realised~\cite{ning2023autonomous}. 
In addition, Li~\emph{et al.} developed a learning-based navigation strategy for transesophageal echocardiography by integrating Vision Transformer with RL~\cite{li2023rl}. All the aforementioned RL approaches in RUSS primarily focus on finding one standard diagnosis plane and achieved robust performance in their specific scenarios. Nonetheless, generating a scanning path to fully cover a specific volume under ribs is yet to be fully investigated. Moreover, compared to the previous RL-based US navigation methods, in this work we try to exploit the usage of CT atlas in US scanning path planning. Such attempt has the potential to convert the partially observable Markov Decision Process to a fully observable one, thus reducing the complexity of the planning task.
% mainly adopted partial observations as state representation, i.e., US frames and action history, which is not directly adaptable to sophisticated cases, e.g., the thoracic applications.

\par
\begin{figure*}[ht!]
\centering
\includegraphics[width=0.90\textwidth]{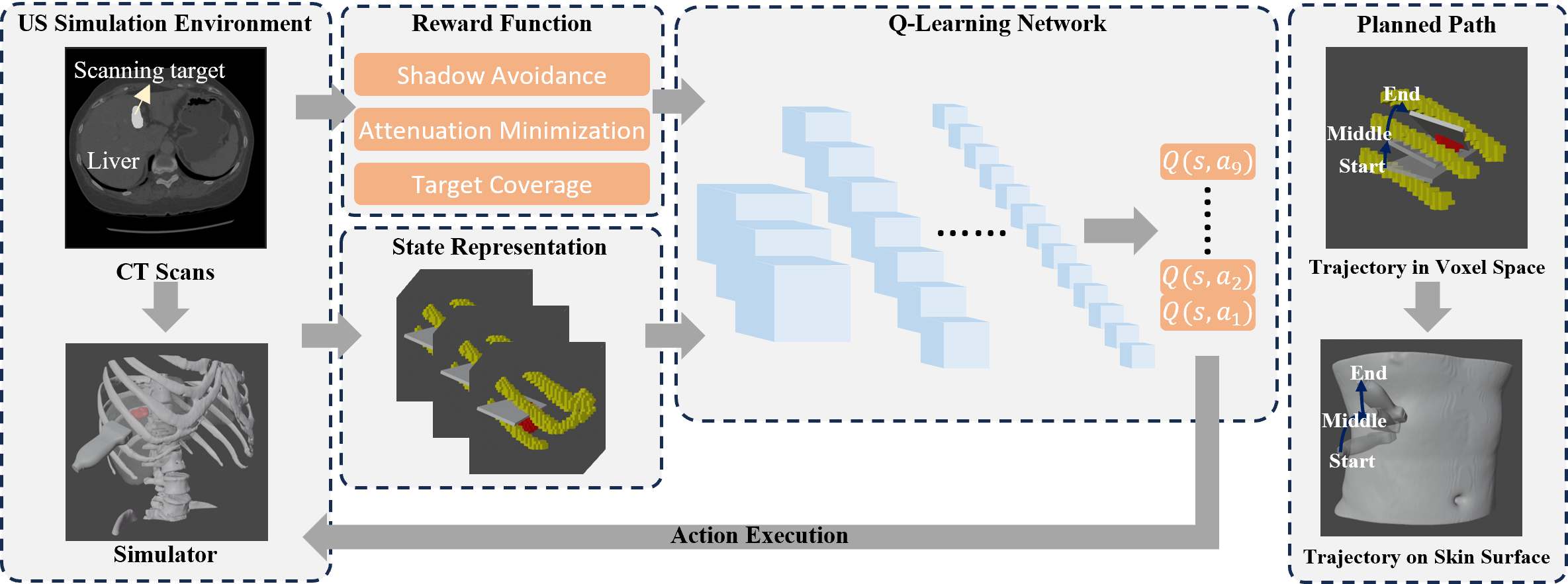}
\caption{Overview of the proposed framework. The scanning target is determined by the doctors on CT and input to a simulator. The overall scanning context is depicted using a 3-channel 3D matrix. Each element within the matrix corresponds to a voxel within the space, and each channel indicates the presence of either a tumor, bone, or an ultrasound ray within the corresponding voxel. Subsequently, an RL model is employed to generate the scanning trajectory. In the end, the planned path is projected back to the skin surface of the patient.%\todo{state representation}
}
\label{Fig_overview}
\end{figure*}

\section{Preliminaries}
\subsection{Deep Q-Learning}
The general goal of an RL agent is to learn a policy $\pi(a_t\mid s_t)$ that maximizes the discounted accumulated rewards $R=\sum_{t=0}^T \lambda^{t}r_t$, where $\lambda$ is the discount factor and $T$ represents the length of one episode. 
Q-function, $Q^{\pi}(s,a)$, is defined as a mapping between state-action pairs and the expected future accumulated reward using policy $\pi$. The optimal Q-function $Q^{\ast}$ can then be defined as the maximum returns given one state-action pair $(s_t,a_t)$ following optimal policy. It satisfies the following Bellman optimization equation $Q^{\ast}(s_t,a_t)=\mathbb{E}[r_t+\lambda \max_{a_{t+1}}Q^{\ast}(s_{t+1},a_{t+1})]$.
% \begin{equation}\label{eq_Qfunction}
% Q^{\ast}(s_t,a_t)=\mathbb{E}[r_t+\lambda \max_{a_{t+1}}Q^{\ast}(s_{t+1},a_{t+1})]
% \end{equation}
% The optimal Q-function can be iteratively updated as shown in Eq.~\ref{eq_bellman}
% \begin{equation}\label{eq_bellman}
% \begin{gathered}
% Q^{\ast}_{new}(s_t,a_t)\leftarrow Q^{\ast}(s_t,a_t)+lr~\left(TD\left(s_t,a_t\right) - Q^{\ast}\left(s_t,a_t\right)\right)\\
% TD(s_t,a_t)=r(s_t,a_t)+\lambda \max_{a_{t+1}}Q^{\ast}(s_{t+1},a_{t+1})
% \end{gathered}
% \end{equation}
% where $TD(s_t,a_t)$ is the temporal difference target and $lr$ denotes the learning rate. 
\par
To deal with complex and large state-action combinations, deep Q-network $Q(s,a \mid \theta)$ is introduced to estimate $Q^{\ast}(s,a)$, where $\theta$ represents the weights of the network~\cite{mnih2013playing}. The loss to update the network is defined as follow:
\begin{equation}\label{eq_loss}
\begin{gathered}
L = \mathbb{E}_{(s_t,a_t,r_t,s_{t+1})}\left[\left(TD(s_t,a_t\mid \theta^-)-Q(s_t,a_t\mid \theta)\right)^2\right]\\
TD(s_t,a_t\mid \theta^-) = r(s_t,a_t)+\lambda \max_{a_{t+1}}Q(s_{t+1},a_{t+1}\mid \theta^-)
\end{gathered}
\end{equation}
where $(s_t,a_t,r_t,s_{t+1})$ are samples drawn from replay buffer, $TD(s_t,a_t)$ is the temporal difference target, and $\theta^-$ denotes the weights in the previous iteration, 
and $Q(s,a\mid \theta^-)$ is also referred as the target network.

\subsection{Double Deep Q-Learning}
The $\max$ operator is used in Eq.~(\ref{eq_loss}) for both action selection and evaluation. This may easily lead to overoptimistic value estimation. To prevent this problem, Hasselt~\emph{et al.} proposed double deep q-learning~\cite{van2016deep}.
The temporal difference target is then modified as follow:
% \begin{multline}
% TD^{DDQL}(s_t,a_t) = r(s_t,a_t)+\\
% \lambda Q\left(s_{t+1},\underbrace{\argmax_{a_{t+1}} Q\left(s_{t+1},a_{t+1}| \theta\right)}_\text{$a_{t+1}$}\middle| ~\theta^-\right)
% \end{multline}
\begin{multline}
TD^{DDQL}(s_t,a_t) = r(s_t,a_t)+\\
\lambda Q\left(s_{t+1},\argmax_{a_{t+1}} Q\left(s_{t+1},a_{t+1}| \theta\right)\middle| ~\theta^-\right)
\end{multline}
where the current network $Q(s,a\mid\theta)$ and the target network $Q(s,a\mid\theta^-)$ are used for action selection and evaluation, respectively.

\subsection{Dueling Deep Q-Learning}
Based on the fact that in many states it is unnecessary to calculate all the values for all actions, Wang~\emph{et al.} proposed dueling deep Q-learning, where the Q-value is decomposed into one state value and one state-dependent action advantage value~\cite{wang2016dueling}. It is realised by separating the deep Q-network into two branches. 
\begin{equation}
Q(s,a\mid \theta, \theta^v, \theta^a)=V(s\mid \theta, \theta^v)+A(s,a\mid \theta, \theta^a)
\end{equation}
where $\theta$, $\theta^v$, and $\theta^a$ are weights of the main network, the state value estimation network, and the advantage value estimation network, respectively. In this work, the dueling deep Q-learning is implemented as the RL framework.

%%%%%%%%%%%%%%%%%%%%%%%%%%%%%%%%%%%%%%%%%%%%%%%%%%%%%%%%%%%%%%%%%%%%%%%%%%%%%%%%%%%%%%%%%%%%%%%%%%%%%%%%%%%%%%%%%%%%%%%%%%%%%%
% \revision{\section{Problem Statement}\label{sec:problemStatement}
% A comprehensive autonomous US scanning system for liver should consist of different components: 1) a path planning module, 2) a registration module, 3) an execution module, and 4) an adjusting module. The planned path is registered to the patient on site and the robot is employed to execute the scanning while establishing proper contact and force with the surface. Considering the dynamical movements of the liver caused by breathing, an adjusting module based on the acquired US images should also be applied to compensate for such variations. This work, as a fundamental component within the broader system, mainly focuses on the design of a robust path planning algorithms tailored to the comprehensive framework.}

%%%%%%%%%%%%%%%%%%%%%%%%%%%%%%%%%%%%%%%%%%%%%%%%%%%%%%%%%%%%%%%%%%%%%%%%%%%%%%%%%%%%%%%%%%%%%%%%%%%%%%%%%%%%%%%%%%%%%%%%%%%%%%

\section{Intercostal Path Planning for Robotic US
% RL Framework for Acoustic Window Planning
}
% Due to the inherited limitation in terms of memory and manipulation accuracy, it is challenging for human operators to maneuver a US probe to image the internal regions of interest completely (3D) through the small intercostal space, even for senior sonographers. 
In order to plan a US scanning path in intercostal spaces to properly visualize internal regions of interest completely, sonographers heavily rely on prior knowledge of 3D anatomical information. Practically, we consider that it is challenging to solely use 2D US images for planning scanning path, given that US often suffers from noise, speckle, and low contrast.
% As discussed above, without having an anatomical model of the human ribs in mind, it is even an unrealistic navigation task for human operators to fully cover a specific volume under acoustic obstacles, while guaranteeing a good image quality.
Therefore, in this study, we train an RL agent based on CT volumes from different patients, providing a 3D understanding of the clinical scenarios. The reasons for not using compounded US volumes are listed as follows: 1) it is hard to reconstruct a comprehensive and correct compounding volume with acoustic obstacles; 2) publicly available US volumes are very rare; 3) the image quality of US images varies a lot between different US systems. For these reasons, CT scans are utilized here to generate 3D scenery views of the rib cage and the scanning target beneath it. The trained model is then used to generate scanning trajectory for unseen patient based on patient-specific CTs. If there is no patient-specific CT available, the path planning can be done on a CT atlas and then the path is registered to the patient on-site through non-regid registration approaches, e.g. the one specific for thoracic applications~\cite{jiang2023thoracic}
% For further steps to mapping the scanning path to patients, a set of registration-based methods~\cite{hennersperger2016towards}, including the one specific for thoracic applications~\cite{jiang2023skeleton}, can be employed. 

\subsection{Environment Setup}\label{sec_env}
The representative target application of this work is intercostal US scans of the right lobe of the liver. Main components of the simulation environment are one CT volume and a virtual US probe. The CT volume serves as an anatomical representation of humans. Based on the given segmentation~\cite{soler20103d}, only the skin surface, rib cage, and target scanning volume of the liver are kept in the simulation environment. \revision{The skin surface is utilized to determine the probe position, while the rib cage and the scanning target are the two main components to assess the visibility of the region of interest in US images.} It is noteworthy that we do not necessarily need a full segmentation of CT. Instead, only the segmentation of bones and skin surface are required, which is relatively easy to achieve, e.g., through thresholding and filtering~\cite{staal2007automatic}. Besides, only the part between the seventh thoracic vertebrae and the fourth lumbar vertebrae is preserved based on the anatomical atlases. In addition, a virtual linear US probe is modeled, where each element on the virtual probe is able to emit a virtual US wave. The propagation process of the wave is simplified as a ray that can go through soft tissues but be blocked by the bones\revision{~\cite{salehi2015patient}}. The direction of the emitted ray is determined by the vertical axis of the probe. \revision{It is noteworthy that no simulated US image is explicitly involved in the proposed path planning pipeline. Rather, it is implicitly represented in the 3D scenery view by visualizing the US imaging plane.}

% \todo{move to actions}
% \begin{figure}[ht!]
% \centering
% \includegraphics[width=0.40\textwidth]{images/cylindrical_coordinate.png}
% \caption{Illustrations of the cylindrical coordinate system.
% }
% \label{Fig_voxelization}
% \end{figure}

% The movement of the virtual probe is constraint on the skin surface to ensure full contact of the probe. A cylindrical coordinate system is built to represent the position of the US probe by fitting a minimum bounding cylinder to the rib cage. Since the movement is constraint on the skin surface, the virtual probe only possesses two translational degrees of freedom (DoF), which can be represented by the height ($h$) and angle ($\theta$) in the cylindrical coordinate system. Considering the inter-patient size variance of the rib cages, the cylinders are affinely registered to the same size. The virtual probe is further projected to the skin surface along the vertical axis of the probe to determine the third translational DoF.

\subsection{State Design}
In general, the true state of the US probe navigation task is the relative pose between the US probe, the anatomy of interest, as well as the acoustic obstacles. In order to compress all these information into one state representation, CT scans of the patients are utilized. It is notable that the CT scans utilized here for training are not patient-specific. Hence, the training CTs can be any publicly available ones that hold comprehensive anatomical representatives. 
\begin{figure}[h!]
\centering
\includegraphics[width=0.48\textwidth]{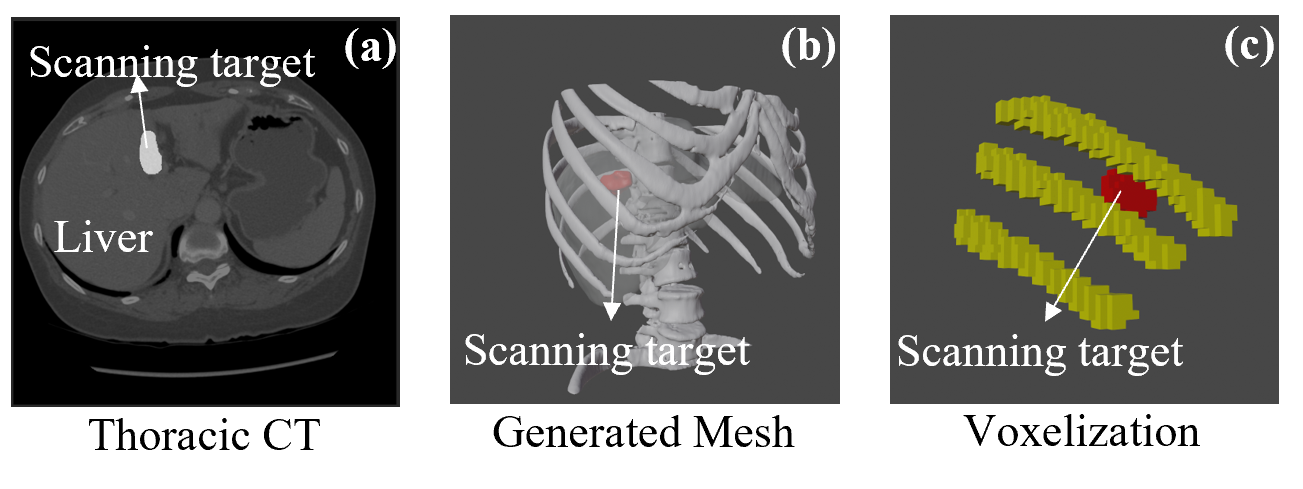}
\caption{Illustrations of the voxelization process.
}
\label{Fig_voxelization}
\end{figure}

A representative thoracic CT is shown in Fig.~\ref{Fig_voxelization}(a). Based on the given segmentation labels, the rib cage and scanning target is transferred to mesh in 3D Slicer\footnote{https://www.slicer.org/}.
% As shown in Fig.~\ref{Fig_voxelization}, the rib cage and the scanning target is segmented from CT scans. 
The shape of the scanning target here is determined by the tumors segmented from CTs or it can also be defined by the doctors manually for scanning specific regions.
Then a surrounding area of $120mm \times120mm\times120mm$ around the scanning target are voxelized by a resolution of $4$ mm in Blender\footnote{https://www.blender.org/}. %\todo{how to select voxelization region} 
,where each voxel has $3$ channels defined as Eq.~(\ref{eq_channels})

\begin{equation}\label{eq_channels}
\begin{array}{l}
F[i,j,k,0] = \left\{ \begin{array}{rl}
1 & target~volume~exists~in~(i,j,k)\\
0 & otherwise \\
\end{array}\right.
\\
F[i,j,k,1] = \left\{ \begin{array}{rl}
1 & bone~exists~in~(i,j,k)\\
0 & otherwise \\
\end{array}\right.
\\
F[i,j,k,2] = \left\{ \begin{array}{rl}
1 & US~rays~pass~through~(i,j,k)\\
\multirow{2}{*}{0} & Bone~occlusion~occurs~or\\
  & out~of~the~imaging~plane \\
\end{array}\right.
\end{array}
\end{equation}
The first channel links to the volume of interest, while the second and third channels indicate the situation of bone structures and mimicked US imaging plane, respectively. Notably, as depicted in Sec.~\ref{sec_env}, the simulated US ray is blocked by the bones, thus generating shadows on the US imaging plane. In order to exploit the time serial information, three consecutive voxelized volumes are jointly used as the state representation in this study. Such a design of state representation is able to integrate all the necessary information in a memory-efficient way compared to using meshes directly.

\subsection{Action Design}\label{sec:action}
\begin{figure}[ht!]
\centering
\includegraphics[width=0.48\textwidth]{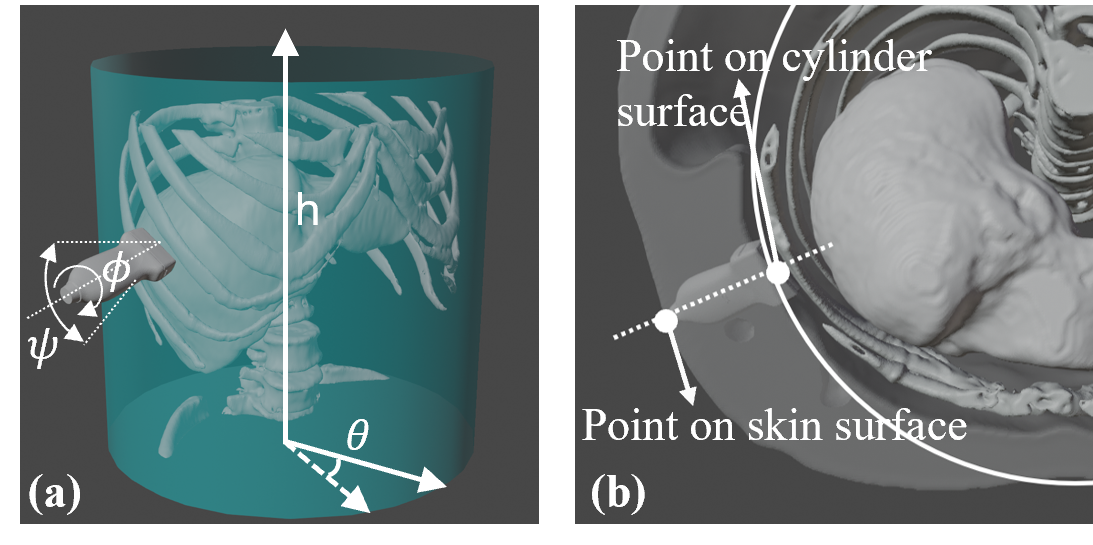}
\caption{(a) Illustrations of the cylindrical coordinate system. (b) Mapping from the cylinder surface to the skin surface. 
}
\label{Fig_cylinder}
\end{figure}

The movement of the virtual probe is constrained on the skin surface to ensure full contact of the probe. A cylindrical coordinate system is built to represent the position of the US probe by fitting a minimum bounding cylinder to the rib cage [Fig.~\ref{Fig_cylinder}(a)]. Since the movement is constrained on the skin surface, the virtual probe only possesses two translational degrees of freedom (DoF), which can be represented by the height ($h$) and angle ($\theta$) in the cylindrical coordinate system. Considering the inter-patient size variance of the rib cages, the cylinders are affinely registered to the same size during training and inference. Later, an inverse resizing process is carried out to map the planned trajectory back to its original scale. The virtual probe is further projected to the skin surface along the vertical axis of the probe to determine the third translational DoF [Fig.~\ref{Fig_cylinder}(b)].

The action space of the proposed RL agent is discrete with four DoFs. Two translational DoFs perpendicular to the probe centerline are realized by moving the probe along the cylinder surface using paired parameters of ($h$,$\theta$) [see Fig.~\ref{Fig_cylinder}]. The step size for $h$ and $\theta$ are empirically set to $4~mm$ and $3^{\circ}$, respectively.   
% and then the movements along the center line of the cylinder with a step of $4$ mm and curved movements around the center line parameterised by $\theta$ with a step size of $3^\circ$. 
The remaining two rotational DoFs are rotations around the probe centerline ($\phi$) and the long axis of the probe footprint (\revision{tilting angle} $\psi$) with the same step size of $2^\circ$. Since the rotations around the short axis of the probe footprint can be replaced by in-plane translations, it is not included in the action space. To ensure that the selected probe pose can be achieved properly in the real scenario, if the angle between the probe centerline and the normal direction of the skin surface is larger than $20^\circ$, the training process will be terminated.

\begin{figure*}[ht!]
\centering
\includegraphics[width=0.85\textwidth]{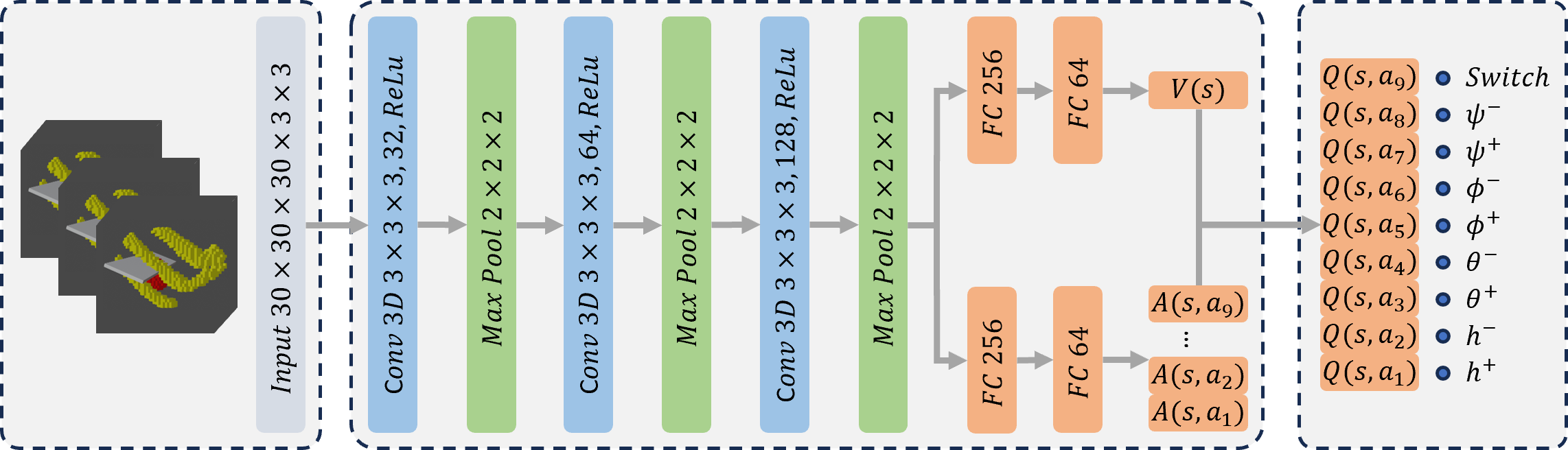}
\caption{Structure of the dueling deep Q-learning network, where $h$ represents the translational movements along the center line of the cylinder coordinate system, $\theta$ denotes the rotation around the cylinder center line, $phi$ is the rotation angle around z-axis of the probe, and $psi$ represents the rotation around the long axis of the probe footprint. $^+$ and $^-$ describe the direction of the actions.
}
\label{Fig_network}
\end{figure*}

\par
Besides the movements in these four DoFs, one extra ``switch" movement is designed to switch the agent between readjusting mode and examining mode. This is motivated by the fact that the target volume may not be properly and fully visualized in a single intercostal gap. Therefore the probe needs to be relocated to neighboring intercostal regions to provide full view of the target anatomy. 
% readjusted to another intercostal region. Therefore, 
In this study, the US images obtained under readjusting mode will not be used in the 3D reconstruction of the target volume. The intuition behind this is to train the agent to have an idle mode allowing it to search from multiple intercostal gaps to fully and effectively visualize the objects. 
% distinguish appropriate imaging poses without acoustic shadows.
% \todo{find a replacement name for adjusting mode}

\subsection{Reward Design}
The reward is designed to encourage the agent to fully cover the target volume as fast as possible, while avoiding the intersection between the US imaging plane and the bones. To realize this, the reward function is decoupled into three aspects.

\subsubsection{Coverage Level of Objects of Interest} The agent is stimulated to cover as much target volume as possible in one sweep step. This reward part represents the most fundamental demand. The reward is then defined as Eq.~(\ref{eq_reward_coverage}).
\begin{equation}\label{eq_reward_coverage}
r_{c} = \frac{n_t}{N}\\
\end{equation}
where $N$ stands for the total number of voxels of the defined regions of interest, and $n_t$ represents the scanned target volume in voxel at step $t$. Notably, if the voxels have already been covered in previous steps, they will not be counted as part of $n_t$. Such design encourages the agent to scan more uncovered regions and converge faster.
\subsubsection{Attenuation Minimization} Attenuation is another unavoidable characteristic of US, which is mainly related to the travelled distance of the emitted US signal\revision{~\cite{salehi2015patient}}. In order to minimize the effect of attenuation on the target volume, a reward term is defined as Eq.~(\ref{eq_reward_attenuation})
\begin{equation}\label{eq_reward_attenuation}
\revision{r_{a} = e^{-\frac{d_t}{R_c}}}\\
\end{equation}
where $d_t$ is the distance between the probe and the scanning target at step $t$, and $R_c$ refers to the radius of the cylinder described in Sec.~\ref{sec:action}. This term is mainly designed to reduce the distance between the probe and the scanning target during path planning so that the agent always tries to image the target from the closest intercostal area.
% $r_c$ refers to the radius of the cylinder described in Sec.~\ref{sec:action}, $D_t$ is then the normalized distance between the target and the probe weighted by $k_a$. In our setup, the scanning target (predefined regions in the right lobe of the liver) is located close to the upper surface of the chest. To obtain relatively large numerical changes of $D_t$ in each step, $k_a$ is set to $4$. 
% erefore, to ensure the numerical discrimination, $k_a$ is set to $4$.

\subsubsection{Shadow Avoidance} Apart from the coverage requirement, in the case of intercostal US scanning, another essential aspect that needs to be properly addressed is the shadow avoidance during the whole scanning process. The shadow cast by bones will impede the interpretability of US images. As a result, a term is defined in the reward function to prevent the agent from acquiring shadow US images.
\begin{equation}\label{eq_reward_shadow}
r_{s} = 1-p_t = 1-\frac{n_{shadow}^t}{N_t}\\
\end{equation}
where $N_t$ refers to the total scanned volume in voxel at step $t$, and $n_{shadow}^t$ is the shadow volume in voxel at step $t$. The shadow volume is defined as the volume, where the simulated US rays are blocked by the bones. As shown in Eq.~(\ref{eq_channels}), the mimicked US imaging plane is not always a complete rectangle. When the occlusion of bones occurs, shadows can be observed on the imaging plane. Therefore, $p_t$ represents the percentage of shadow areas in the volume covered by the probe at step $t$. By negatively correlating the size of the shadow volume with the reward term defined in Eq.~(\ref{eq_reward_shadow}), the agent is punished when shadow occurs.

\par
% In order to stimulate the agent to scan over more volume of the target in one step, the agent is give a positive reward based on the number of voxels scanned on the target $n_t$ in current step. By correlating the reward with the inverse of the average distance between probe and scanning target $d_t$ during one step, the agent is encouraged to find a path where the probe is closer to the target. Thus, the attenuation of the US wave is minimized to have a better visualization. Meanwhile, to avoid the agent from taking images with shadows created by bones, $p_t$ is defined as the percentage of non-shadow areas in the volume covered by the probe in the current step. 
Combining all three terms, the reward is defined as follows:
\begin{equation}\label{eq_reward_step}
r_t = r_{c}+\alpha_1 r_{a} + \alpha_2 r_{s}
\end{equation}
where $\alpha_1,~\alpha_2\in [0,1]$ are hyperparameters used to balance the effects caused by the three terms. Since the first term is closely related to the fundamental fulfillment of the objective task, it is always kept as $1$.
Once $95\%$ of the whole target volume has been obtained, the path planning is considered as success, and a high reward value will be assigned as Eq.~(\ref{eq_reward_final}).

\begin{equation}\label{eq_reward_final}
\begin{gathered}
r_{end} = k_{end}~(1+\alpha_1 D + \alpha_2 P)\\
D = \frac{1}{T}\sum_{t=1}^{T} \frac{d_t}{R_c},~ P = \frac{1}{T}\sum_{t=1}^{T} (1-p_t)
\end{gathered}
\end{equation}
where $P$ represents the average non-shadow percentage during the scanning process, $D$ refers to the average value of the normalized distance between probe and scanning target, and $k_{end}$ is a coefficient to grant the agent a high reward after a successful completion.

It needs to be noted that there will be no punishment or reward accumulated under readjusting mode as depicted in Sec.~\ref{sec:action}. However, the switch action to de- or activate the readjusting mode will result in a one-time negative reward to prevent redundant switching actions.
In order to force the agent to acquire shadow-free images during the scanning, a threshold value $T_{th}$ on $p_t$ is introduced. Only if the percentage of shadow volume $p_t$ is below $T_{th}$ at the current step, augmented voxels of the target are counted and a positive reward is awarded. When $p_t$ surpasses $T_{th}$, even if the agent is able to cover a portion of the target volume, the volume will not be counted and a negative reward is assigned. By doing so, the model is trained to prevent shadows during path planning. 
% Meanwhile, a negative reward is given when $p_t$ in the current step is lower than a threshold, indicating that only a small portion of the scanned volume in this step is shadow-free.
% If the current trial is able to generate trajectory with higher average percentage of target volume in each step and lower overall distance between the probe and the scanning target, then a higher reward is assigned.
The final reward function is defined in Eq.~(\ref{eq_rewards}).
\begin{equation}\label{eq_rewards}
r = \left\{ \begin{array}{cl}
r_t & p_{t}<T_{th},adj=False\\
r_{end} & \sum n_t>0.95N,adj=False\\
-1 & a_{t}=switch\\
-0.1 & p_t \geq T_{th}, adj=False\\
0 & adj=True
\end{array}\right.
\end{equation}
$\sum n_t$ stands for the accumulated number of scanned voxels of the target, $adj$ represents whether the readjusting mode is activated, and $a_t$ is the current action. 

\subsection{Network Design}
The double dueling deep Q-learning is selected as the training pipeline. The network structure is shown in Fig.~\ref{Fig_network}. The network is consisted of consecutive 3D convolutional layers with ReLU as activation function followed by max pooling layers. At the end of the convolutional process, the feature maps are flattered and fed into two streams of fully connected layers with ReLU to estimate the value function at the given state and the advantage function of all possible actions. By combining the estimations of the two branches, the Q-value function is calculated following Eq.~(\ref{eq_qvalue}) to increase the stability during the training~\cite{wang2016dueling}. 
\begin{multline}\label{eq_qvalue}
Q(s,a_t\mid \theta,\theta^v,\theta^a)=V(s\mid \theta^v)+\\
\left(A(s,a_t\mid \theta^a)-\frac{1}{N_a}\sum_{t=1}^{N_a} A(s,a_t\mid \theta^a)\right)
\end{multline}
where $N_a$ is the dimension of the discrete action space.
% \begin{figure*}[ht!]
% \centering
% \includegraphics[width=0.85\textwidth]{images/network_structure.png}
% \caption{Structure of the dueling deep Q-learning network, where $h$ represents the translational movements along the center line of the cylinder coordinate system, $\theta$ denotes the rotation around the cylinder center line, $phi$ is the rotation angle around z-axis of the probe, and $psi$ represents the rotation around the long axis of the probe footprint. $^+$ and $^-$ describe the direction of the actions.}
% \label{Fig_network}
% \end{figure*}

\begin{table*}[ht!]
\caption{Results of Ablation study using Varying hyperparameter}\label{tab:ablation}\centering
\label{tab:th_table}
\resizebox{\textwidth}{!}{
  \begin{tabular}{c c c c| c| c| c| c| c| c| c| c| c| c| c}
    \toprule
    \multicolumn{3}{c}{Hyperparameter}&\multicolumn{3}{c|}{Success rate}&\multicolumn{3}{c|}{Average steps}&\multicolumn{3}{c|}{Average P ($\%$)}&\multicolumn{3}{c}{Average D ($\%$)}\\
    \cmidrule{4-15}
    $T_{th}$ & $\alpha_1$ & $\alpha_2$ & S & M & L & S & M & L & S & M & L & S & M & L\\

    \midrule
    $0$ & $1$ & $0.5$     & $88\%$ & $94\%$ & $76\%$ & $14.7 (7.9)$ & $23.9 (9.6)$ & $40.0 (13.7)$ & $58.0 (1.7)$ & $60.7 (2.1)$ &$58.7 (0.9)$ & $35.6 (0.4)$ & $36.0 (0.2)$ & $36.4 (0.3)$ \\
    $0.8$ & $1$ & $0.5$   & $95\%$ & $92\%$ & $81\%$ & $25.3 (8.9)$ & $31.4 (10.2)$ & $41.3 (10.2)$ & $92.4 (0.7)$ & $93.3 (0.8)$ &$92.2 (1.3)$ & $35.9 (0.3)$ & $36.2 (0.2)$ & $36.7 (0.3)$ \\
    $0.95$ & $1$ & $0.5$  & $90\%$ & $85\%$ & $61\%$ & $30.5 (7.1)$ & $35.7 (8.6)$ & $47.0 (12.5)$ & $99.5 (0.1)$ & $99.1 (0.2)$ &$99.1 (0.3)$ & $36.0 (0.3)$ & $36.4 (0.2)$ & $37.3 (0.3)$ \\
    $0$ & $0$ & $1$         & $90\%$ & $91\%$ & $76\%$ & $15.7 (8.2)$ & $22.4 (9.4)$ & $36.7 (11.6)$ & $61.0 (1.3)$ & $62.5 (1.7)$ &$60.7 (2.0)$ & $36.2 (0.4)$ & $36.6 (0.4)$ & $36.9 (0.4)$ \\
    $0$ & $2$ & $0$         & $94\%$ & $87\%$ & $80\%$ & $14.3 (7.5)$ & $22.9 (9.3)$ & $37.1 (12.4)$ & $58.7 (1.9)$ & $58.5 (2.3)$ &$60.1 (1.7)$ & $35.3 (0.3)$ & $35.7 (0.3)$ & $35.9 (0.4)$ \\
    % $0$ & $0$ & $0$         & $94\%$ & $92\%$ & $82\%$ & $13.2 \pm 7.8$ & $21.8 \pm 8.9$ & $38.2 \pm 12.0$ & $57.2\pm 1.8\%$ & $58.3\pm 1.8\%$ &$60.1\pm 1.9\%$ & $72.3 \pm 0.7\%$ & $73.5 \pm 0.6\%$ & $73.7 \pm 0.9\%$ \\
    \bottomrule
  \end{tabular}
  }
  \begin{tablenotes}
        \footnotesize
        \item mean(std).
  \end{tablenotes}
\end{table*}

% problem
% env augmentation ribs and targeSkets
% from cylinder to skin surface, orientation between the probe z-axis and the normal direction of surface
% rotation actions dianye
% alpha beta gamma effect
% th effect
% convex?
% cylinder -> ellipse cylinder
% find another name for "adjusting" mode
\section{Experiments}
\subsection{Implementation Details}
% The double dueling deep Q-learning with prioritized replay buffer is selected as the training pipeline. The network structure is shown in Fig.~\ref{Fig_network}. The input volumes are fed into three consecutive 3D-convolutional and max pooling layers to extract meaningful features. Then the extracted feature maps are flattened and led into two separate fully connected layers to estimate the state value and advantage values, respectively. The Q-value is calculated following Eq.~\ref{eq_qvalue} to increase the stability during the training~\cite{wang2016dueling}. 
% \begin{multline}\label{eq_qvalue}
% Q(s,a_t\mid \theta,\theta^v,\theta^a)=V(s\mid \theta^v)+\\
% \left(A(s,a_t\mid \theta^a)-\frac{1}{N_a}\sum_{t=1}^{N_a} A(s,a_t\mid \theta^a)\right)
% \end{multline}
% where $N_a$ is the dimension of the discrete action space. 
The prioritized replay is implemented to realize faster and better learning performance~\cite{schaul2015prioritized}. The learning rate is set to $7\times 10^{-5}$. The maximum number of steps in each trial is $80$. The model is trained for a total number of $5\times 10^6$ steps, while the target network is updated every $5\times 10^3$ steps. The exploration rate decays from $1$ to $0.05$ linearly in the first $3\times 10^6$ steps. 
% The learning rate decays from $1\times 10^{-4}$ with the decay rate of $0.85$ and decay step of $5\times 10^5$.
% Both $\alpha_1$ and $\alpha_2$ were $0.5$. 
To efficiently simulate the extensive training data within a reasonable time period, we employ distributed RL across $16$ training nodes.
% using OpenAI Gym \cite{brockman2016openai} and Tianshou \cite{tianshou}. 
Each node is assigned with a local replay buffer with size of $5\times 10^3$.

\par
To train and validate the method, we manually selected $20$ liver tumors and $8$ rib cages from different CTs from public dataset 3D-IRCADb-01~\cite{soler20103d}. Four tumors and two rib cages are utilized as test data, while the others are employed to build up the training environment.
% The training and testing dataset are divided in ratios of $8:2$ and $3:1$, respectively. 
% to allocate training and test samples. 
To eliminate the size differences between different patients, all rib cages are affinely resized to a generic cylindrical coordinate based on the average size of human bodies in the applied dataset. Afterwards, an inverse resizing process is conducted to map the planned trajectory back to the real scale. The liver tumors in this dataset were employed as scanning targets. 
% The size of tumors are firstly normalized to similar sizes. 
In each episode of training, the position, orientation and size of tumor are randomly initialized in a chosen rib cage in order to enrich the variability of training environment. The training was conducted on a PC with AMD EPYC 24-core CPU and RTX4090 GPU. The virtual environment is visualized in Blender.

\subsection{Ablation Experiments}

In the ablation study, the performance of different models was compared with each other by success rate, average steps, average $P$, and average $D$. If the agent can cover $95\%$ of the target volume within 80 steps, it is considered as a successful case. The average $P$ and $D$ are utilized to show the performance of the models on shadow avoidance and attenuation minimization, respectively. Considering the inter-patient variations, the targets of interest are further categorized into three classes based on their sizes, i.e., small (less than $4cm^3$), medium (ranging from $4cm^3$ to $13.5cm^3$), and large (exceeding $13.5cm^3$). Each ablation case (each row in Table~\ref{tab:ablation}) is evaluated by $100$ experiments, in total 500 experiments. %\todo{number of experiments in each case}
% The performance of each model is then evaluated under each category.

\subsubsection{Threshold Parameter of Shadow Region}
Firstly, the ablation study was conducted to validate the effect of threshold value $T_{th}$ in reward function Eq.~(\ref{eq_rewards}). If the portion of shadow volume $p_t$ at current step is larger than the threshold value $T_{th}$, a negative reward will be assigned to the agent to increase the shadow awareness of the trained agent. The other hyperparameters were set as the default values, i.e., $\alpha_1=1$, and $\alpha_2=0.5$. Three models are trained with different threshold values, i.e., $T_{th}=0$, $T_{th}=0.8$, and $T_{th}=0.95$. 

\par
As shown in the first three rows of Table~\ref{tab:ablation}, the model trained with $T_{th}=0.8$ surpasses the performance of the other two models in most cases across all tumor sizes in terms of success rate. The only exception happens when $T_{th}=0$ for the medium-size tumor ($92\%$ vs $94\%$). 
% except for a lag of 2\% for the medium-size tumor, compared with the model trained with $T_{th}=0$. 
Interestingly, these results are somehow counter-intuitive. 
One would have expected that the agent trained with $T_{th}=0$ to yield the highest success rate since it possesses no constraint at all.
Such intuition holds true during the training phase, the model with $T_{th}=0$ has the highest success rate of $95\%$ on training sets. However, it fails to generalize on unseen tumor shapes and rib cages during the test phase, resulting in a diminished success rate compared to its training performance. If the agent is not sufficiently constrained regarding the scan quality during training, it cannot learn the occlusion relationship between the ribs and tumors efficiently.

\par
When the threshold value increases, the average $P$ of each model also rises since the essence of $T_{th}$ is to impose a hard constraint to prevent the existence of shadows during trajectory planning. Apart from the average percentage of non-shadow volume $P$, the average steps and average distance $D$ between probe and target on cylindrical coordinate also increase under a higher threshold value. With the increasingly severe constraint, it is reasonable for the agent to perform extra steps to find an appropriate acoustic window and visualize the target volume from a distance to avoid intersection with bones.

\subsubsection{Hyperparameters of Attenuation Minimization and Shadow Avoidance}
The reward function consists of three terms: the target coverage term, the attenuation minimization term, and the shadow avoidance term.
% Since the first term is closely related to the fundamental fulfillment of the objective task, it is not practical to eliminate it from the reward function. Therefore, only the coefficients, i.e., $\beta$ and $\gamma$, of the second and third term are manipulated in this section to evaluate the influence of each corresponding term. 
As shown in the last two rows of Table~\ref{tab:ablation}, when the attenuation minimization term ($\alpha_1$) is removed from the reward function, the average $D$ increases in all three different target sizes compared with standard hyperparameter setup ($T_{th}=0$, $\alpha_1=1$, $\alpha_2=0.5$). The increase in average $D$ is even more significant when comparing it with setting $\alpha_1$ to two and setting $\alpha_2$ to zero (fifth row of Table~\ref{tab:ablation}), which demonstrates the effectiveness of explicitly considering the attenuation minimization term ($\alpha_1$) in the reward function. For the case when the shadow avoidance term is restrained, the average portion of non-shadow volume decreases compared to the case when only shadow avoidance term is included. Thus, it can be concluded that the attenuation minimization term and the shadow avoidance term are able to achieve the expected effects, respectively.

%  The reward function is composed of three elements: the coverage term, the non-shadowed region term, and the distance term. We then evaluate the influence of the modification of reward function by eliminating one or both of the last two terms. The first term directs the agent to fulfill the fundamental task. Therefore, we naturally do not eliminate it. The results are shown in Table \ref{tab:w_table}. It is obvious that, the policy achieves a slightly higher average non-shadow area proportion, when $\gamma$ is set to 1. Conversely, it achieves a slightly lower average distance, when $\beta$ is set to 1. They indicate that both of last two terms in reward function are actively influencing the agent's behavior.

\begin{table*}
\centering
\begin{threeparttable}
\caption{Performance of the proposed RL model on multiple targets scanning task.}\label{tab:multi}
\label{tab:n_table}
  \begin{tabular}{C{0.08\textwidth} C{0.065\textwidth}| C{0.065\textwidth}| C{0.091\textwidth}| C{0.091\textwidth}| C{0.091\textwidth}| C{0.091\textwidth}| C{0.091\textwidth}| C{0.091\textwidth}}
    \toprule
    \multirow{2}{*}{\shortstack[c]{Number of\\targets}}
    &\multicolumn{2}{c|}{Success rate}&\multicolumn{2}{c|}{Average steps}&\multicolumn{2}{c|}{Average P ($\%$)}&\multicolumn{2}{c}{Average D ($\%$)}\\
    \cmidrule{2-9}
     & S & M & S & M & S & M & S & M\\

    \midrule
    $1$     & $95\%$ & $92\%$ & $25.3 (6.1)$ & $31.9 (7.5)$ & $95.6 (0.4)$ & $95.2 (0.4)$ & $36.6 (0.4)$ & $36.8 (0.4)$ \\
    $2$     & $91\%$ & $86\%$ & $37.3 (10.9)$ & $44.7 (9.6)$ & $95.2 (0.8)$ & $94.8 (0.9)$ & $37.3 (0.2)$ & $37.5 (0.2)$ \\
    $3$     & $90\%$ & $84\%$ & $42.9 (9.7)$ & $51.4 (10.1)$ & $95.0 (0.7)$ & $94.6 (0.9)$ & $37.5 (0.3)$ & $37.7 (0.3)$ \\
    \bottomrule
  \end{tabular}
  \begin{tablenotes}
        \footnotesize
        \item mean(std).
  \end{tablenotes}
\end{threeparttable}
\end{table*}

\begin{figure}[ht!]
\centering
\includegraphics[width=0.48\textwidth]{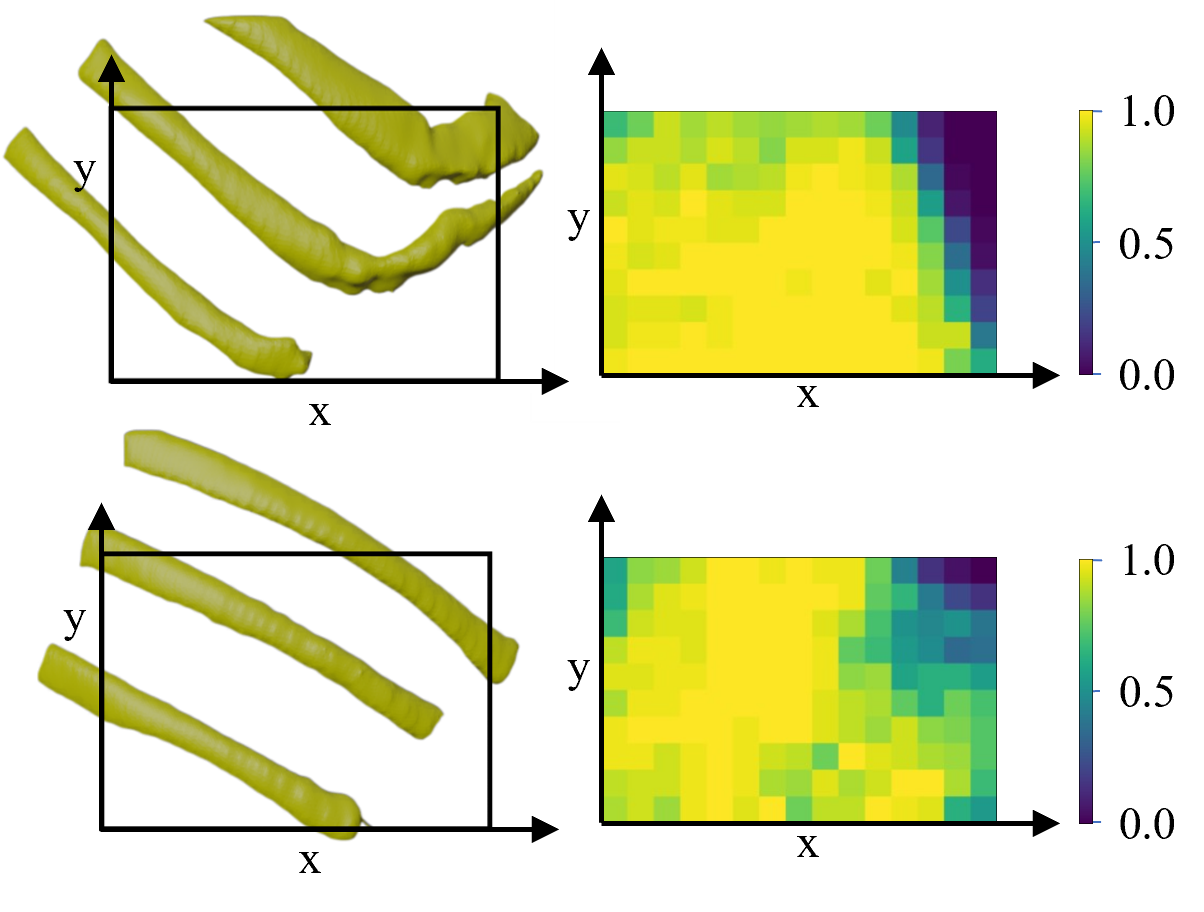}
\caption{Heatmaps of success rates based on scanning target positions in the X-Y plane defined in CT coordinate.}
\label{Fig_heat}
\end{figure}

\subsection{Impact of Target Locations}
% Success Rate Distribution
To investigate the impact of the target positions on success rate, ten segmented tumors are initialized at the center point of every voxel beneath the ribs, where the target tumor has no intersection with the ribs. The success rate is computed for each possible configuration. 
% After the voxelization of the intercostal area, the segmented tumors are positioned at the center point of every voxel beneath the ribs. 
% Then, if the target tumor has no intersection with ribs, this initialization is adopted for validation.
% ; otherwise, the location of the target tumor will be reinitialized. 
% as the initialization position of the ten scanning targets. 
The model trained with the hyperparameter configuration of $T_{th}=0.8$, $\alpha_1=1$, and $\alpha_2=0.5$ was employed on two unseen ribs in this study. 
% For visualization purpose, we randomly initialized $z_{ct}$ value of ten tumors at each combinations of ($x_{ct}$, $y_{ct}$) [see Fig.~\ref{Fig_heat}].
% For each target volume, at every voxel center, ten experiments are carried out to evaluate the impact of tumor location on the proposed method.
% In total, $10$ experiments have been carried out to evaluate the impact of each tumor location on the proposed method. \todo{number of carried out experiments}

\par
% The success rate on x-y plane is displayed in Fig.~\ref{Fig_heat}. Each pixel value represents the averaged success rate over z-axis.
% The success rate at each ($x_{ct}$, $y_{ct}$) pair is shown in Fig.~\ref{Fig_heat}. 
Since the position of the target tumor along the z-axis in CT coordinate (spanning from front to back) does not have a big impact on the success rate compared to x- and y-axis, therefore, the success rates over the z-axis were averaged and displayed on the x-y plane as shown in Fig.~\ref{Fig_heat}. 
It can be seen that regions with narrower intercostal spaces exhibit a relatively lower success rate compared to other areas. For instance, the success rate in the upper right section of the first ribcage is notably low, where the intercostal space between ribs is only $10mm$. This compact spacing poses a significant challenge for the agent to scan these areas effectively while still achieving a relatively high proportion of non-shadowed areas ($T_{th}=0.8$).

% To investigate the impact of tumor position on success rate, we initialize the tumor with different positions and assess the success rate for each configuration. We conduct the evaluation on the two ribs from test dataset with the policy trained with $th=3$ and $\beta=\gamma=0.5$. We find that the tumor's position on the z-axis (spanning from the body's skin to the liver) doesn't significantly affect the success rate, therefore, we average out the success rates over the z-axis and present the results as 2D heatmaps (Fig. \ref{Fig_heat}), focusing solely on the variations along the x and y axes. From the heatmap, we observe that regions with narrower intercostal spaces between the ribs exhibit a relatively lower success rate compared to other areas. For instance, the success rate in the upper right section of the first ribcage is notably low, where the intercostal space between ribs is only $10mm$ there. This tight spacing presents a significant challenge for the agent to scan these areas effectively while still achieving a relatively high proportion of non-shadowed areas ($th=0.8$).

\subsection{Results on Multiple Scanning Targets}
Considering that lesions could be located in various areas, especially if the tumor was incompletely removed after resection. It would also be important to ascertain if the proposed framework can effectively operate in configurations with multiple scanning targets. Therefore, we randomized the number of targets from 1 to 3 and retrained a policy for scanning multiple targets. To reduce the probability of targets overlapping, the large-sized target was not used in this validation. During the training, the number of targets was randomly selected at initialization. The hyperparameter configuration adopted here was $T_{th}=0.8$, $\alpha_1=1$, and $\alpha_2=0.5$ as well.
The results are shown in Table \ref{tab:multi}. The success rate drops as the number of targets increases. However, the trained model can still deliver decent success rates for both small- and medium-sized targets with $90\%$ and $84\%$ respectively. As expected, the average steps and average distance between the probe and the target also raised with the increasing number of scanning targets. We can conclude that the proposed framework is able to plan the trajectory to cover multiple scanning targets at the same time.
% We find that even in scenarios with multiple tumors, our framework consistently attains a success rate of a minimum of $90\%$ for small-sized tumors and $84\%$ for medium-sized tumors.
% It would also be intriguing to ascertain if the proposed framework can effectively operate in configurations with multiple tumors. Therefore, we randomize the number of tumors from 1 to 3 and additionally train a policy under this condition. To reduce the probability of tumors overlapping, we omit the large-sized tumors in this setup. The results are shown in Table \ref{tab:n_table}. We find that even in scenarios with multiple tumors, our framework consistently attains a success rate of a minimum of $90\%$ for small-sized tumors and $84\%$ for medium-sized tumors.

\section{Conclusion and Discussions}
In this work, we present an RL approach to automatically generate a US scanning path aiming to fully cover and reconstruct single or multiple arbitrary target volumes in intercostal spaces, while minimizing overall acoustic attenuation and shadow. The training is conducted in simulation environment using CT atlas. Compared to US image based navigation, utilizing directly the anatomical models as state representation benefits the training in the sense that the whole navigation process is not partially but fully observable. The ablation study has validated the effectiveness of the proposed framework (success rate: $95\%$, $92\%$, and $81\%$ for unseen small-, medium-, and large-sized target; success rate: $90\%$ and $84\%$ for three small- and medium-sized targets).
\par
This work marks the initial step in developing a fully autonomous US scanning system for intercostal applications. 
A comprehensive autonomous US scanning system for liver should consist of different components, e.g., path planning module, registration module, and robot control module. The planned scanning path is registered to the patient on-site and the robot is employed to execute the scanning while ensuring appropriate contact and force interaction with the surface. By applying non-rigid registration approaches~\cite{jiang2023skeleton,jiang2023thoracic}, we can then plan the trajectory on a CT atlas and realize non-patient-specific path projection.
One possible application scenario for such robotic system is liver ablations. The robot can then help the surgeons to monitor the target tumor as a whole during and after the ablation to assess the surgical outcomes.
This work, as a fundamental component within the broader system, is primarily devoted to designing a robust path planning algorithms tailored to the comprehensive framework.
% The current work mainly focuses on the path planning module of the border system.
% In future work, the feasibility of applying skeleton-based registration method~\cite{jiang2023skeleton} to map the pre-planned trajectory to the patient on-site could be evaluated.}
% , achieving $95\%$, $92\%$, and $81\%$ success rate on unseen small-, medium-, and large-sized target. 
% In future work, the generated trajectory can be further registered to the patient on-site by applying skeleton-based registration method~\cite{jiang2023skeleton}.
In addition, future studies will further validate the performance of the proposed method on volunteers or patients in real-world scenarios by considering more practical factors, such as image quality and US probe contact conditions.

\bibliographystyle{IEEEtran}
\bibliography{IEEEabrv,references}

\begin{thebibliography}{10}
\providecommand{\url}[1]{#1}
\csname url@samestyle\endcsname
\providecommand{\newblock}{\relax}
\providecommand{\bibinfo}[2]{#2}
\providecommand{\BIBentrySTDinterwordspacing}{\spaceskip=0pt\relax}
\providecommand{\BIBentryALTinterwordstretchfactor}{4}
\providecommand{\BIBentryALTinterwordspacing}{\spaceskip=\fontdimen2\font plus
\BIBentryALTinterwordstretchfactor\fontdimen3\font minus \fontdimen4\font\relax}
\providecommand{\BIBforeignlanguage}[2]{{%
\expandafter\ifx\csname l@#1\endcsname\relax
\typeout{** WARNING: IEEEtran.bst: No hyphenation pattern has been}%
\typeout{** loaded for the language `#1'. Using the pattern for}%
\typeout{** the default language instead.}%
\else
\language=\csname l@#1\endcsname
\fi
#2}}
\providecommand{\BIBdecl}{\relax}
\BIBdecl

\bibitem{fiorentino2022deep}
M.~C. Fiorentino, E.~Cipolletta, E.~Filippucci, W.~Grassi, E.~Frontoni, and S.~Moccia, ``A deep-learning framework for metacarpal-head cartilage-thickness estimation in ultrasound rheumatological images,'' \emph{Computers in Biology and Medicine}, vol. 141, p. 105117, 2022.

\bibitem{mustafa2013development}
A.~S.~B. Mustafa, T.~Ishii, Y.~Matsunaga, R.~Nakadate, H.~Ishii, K.~Ogawa, A.~Saito, M.~Sugawara, K.~Niki, and A.~Takanishi, ``Development of robotic system for autonomous liver screening using ultrasound scanning device,'' in \emph{2013 ROBIO}.\hskip 1em plus 0.5em minus 0.4em\relax IEEE, 2013, pp. 804--809.

\bibitem{jiang2022towards}
Z.~Jiang, Y.~Gao, L.~Xie, and N.~Navab, ``Towards autonomous atlas-based ultrasound acquisitions in presence of articulated motion,'' \emph{IEEE Robotics and Automation Letters}, vol.~7, no.~3, pp. 7423--7430, 2022.

\bibitem{ipsen2021towards}
S.~Ipsen, D.~Wulff, I.~Kuhlemann, A.~Schweikard, and F.~Ernst, ``Towards automated ultrasound imaging—robotic image acquisition in liver and prostate for long-term motion monitoring,'' \emph{Physics in Medicine \& Biology}, vol.~66, no.~9, p. 094002, 2021.

\bibitem{si2024design}
W.~Si, N.~Wang, and C.~Yang, ``Design and quantitative assessment of teleoperation-based human--robot collaboration method for robot-assisted sonography,'' \emph{IEEE Trans. Autom. Sci. Eng.}, 2024.

\bibitem{dyck2024towards}
M.~Dyck, A.~Weld, J.~Klodmann, A.~Kirst, L.~Dixon, G.~Anichini, S.~Camp, A.~Albu-Sch{\"a}ffer, and S.~Giannarou, ``Towards safe and collaborative robotic ultrasound tissue scanning in neurosurgery,'' \emph{IEEE Transactions on Medical Robotics and Bionics}, 2024.

\bibitem{jiang2023robotic}
Z.~Jiang, S.~E. Salcudean, and N.~Navab, ``Robotic ultrasound imaging: State-of-the-art and future perspectives,'' \emph{Medical image analysis}, p. 102878, 2023.

\bibitem{li2021overview}
K.~Li, Y.~Xu, and M.~Q.-H. Meng, ``An overview of systems and techniques for autonomous robotic ultrasound acquisitions,'' \emph{IEEE Transactions on Medical Robotics and Bionics}, vol.~3, no.~2, pp. 510--524, 2021.

\bibitem{jiang2020automatic}
Z.~Jiang, M.~Grimm, M.~Zhou, Y.~Hu, J.~Esteban, and N.~Navab, ``Automatic force-based probe positioning for precise robotic ultrasound acquisition,'' \emph{IEEE Transactions on Industrial Electronics}, vol.~68, no.~11, pp. 11\,200--11\,211, 2020.

\bibitem{akbari2021robot}
M.~Akbari, J.~Carriere, R.~Sloboda, T.~Meyer, N.~Usmani, S.~Husain, and M.~Tavakoli, ``Robot-assisted breast ultrasound scanning using geometrical analysis of the seroma and image segmentation,'' in \emph{2021 IROS}.\hskip 1em plus 0.5em minus 0.4em\relax IEEE, 2021, pp. 3784--3791.

\bibitem{huang2024robot}
Q.~Huang, B.~Gao, and M.~Wang, ``Robot-assisted autonomous ultrasound imaging for carotid artery,'' \emph{IEEE Transactions on Instrumentation and Measurement}, 2024.

\bibitem{huang2021towards}
Y.~Huang, W.~Xiao, C.~Wang, H.~Liu, R.~Huang, and Z.~Sun, ``Towards fully autonomous ultrasound scanning robot with imitation learning based on clinical protocols,'' \emph{IEEE Robotics and Automation Letters}, vol.~6, no.~2, pp. 3671--3678, 2021.

\bibitem{bal2023curvature}
A.~Bal, A.~Gupta, F.~Abhimanyu, J.~Galeotti, and H.~Choset, ``A curvature and trajectory optimization-based 3d surface reconstruction pipeline for ultrasound trajectory generation,'' in \emph{ICRA}.\hskip 1em plus 0.5em minus 0.4em\relax IEEE, 2023, pp. 2724--2730.

\bibitem{tan2023autonomous}
J.~Tan, J.~Li, Y.~Li, B.~Li, Y.~Leng, Y.~Rong, and C.~Fu, ``Autonomous trajectory planning for ultrasound-guided real-time tracking of suspicious breast tumor targets,'' \emph{IEEE Trans. Autom. Sci. Eng.}, 2023.

\bibitem{gobl2017acoustic}
R.~G{\"o}bl, S.~Virga, J.~Rackerseder, B.~Frisch, N.~Navab, and C.~Hennersperger, ``Acoustic window planning for ultrasound acquisition,'' \emph{International Journal of Computer Assisted Radiology and Surgery}, vol.~12, pp. 993--1001, 2017.

\bibitem{li2021image}
K.~Li, Y.~Xu, J.~Wang, D.~Ni, L.~Liu, and M.~Q.-H. Meng, ``Image-guided navigation of a robotic ultrasound probe for autonomous spinal sonography using a shadow-aware dual-agent framework,'' \emph{IEEE Trans. Med. Robot. Bionics}, vol.~4, no.~1, pp. 130--144, 2021.

\bibitem{bi2022vesnet}
Y.~Bi, Z.~Jiang, Y.~Gao, T.~Wendler, A.~Karlas, and N.~Navab, ``Vesnet-rl: Simulation-based reinforcement learning for real-world us probe navigation,'' \emph{IEEE Robotics and Automation Letters}, vol.~7, no.~3, pp. 6638--6645, 2022.

\bibitem{hase2020ultrasound}
H.~Hase, M.~F. Azampour, M.~Tirindelli, M.~Paschali, W.~Simson, E.~Fatemizadeh, and N.~Navab, ``Ultrasound-guided robotic navigation with deep reinforcement learning,'' in \emph{2020 IROS}.\hskip 1em plus 0.5em minus 0.4em\relax IEEE, 2020, pp. 5534--5541.

\bibitem{deng2021learning}
X.~Deng, Y.~Chen, F.~Chen, and M.~Li, ``Learning robotic ultrasound scanning skills via human demonstrations and guided explorations,'' in \emph{ROBIO}.\hskip 1em plus 0.5em minus 0.4em\relax IEEE, 2021, pp. 372--378.

\bibitem{soler20103d}
L.~Soler, A.~Hostettler, V.~Agnus, A.~Charnoz, J.~Fasquel, J.~Moreau, A.~Osswald, M.~Bouhadjar, and J.~Marescaux, ``3d image reconstruction for comparison of algorithm database: A patient specific anatomical and medical image database,'' \emph{IRCAD, Strasbourg, France, Tech. Rep}, vol.~1, no.~1, 2010.

\bibitem{jiang2023skeleton}
Z.~Jiang, X.~Li, C.~Zhang, Y.~Bi, W.~Stechele, and N.~Navab, ``Skeleton graph-based ultrasound-ct non-rigid registration,'' \emph{IEEE Robotics and Automation Letters}, 2023.

\bibitem{jiang2023thoracic}
Z.~Jiang, C.~Li, X.~Lil, and N.~Navab, ``Thoracic cartilage ultrasound-ct registration using dense skeleton graph,'' in \emph{2023 IEEE/RSJ International Conference on Intelligent Robots and Systems (IROS)}.\hskip 1em plus 0.5em minus 0.4em\relax IEEE, 2023, pp. 6586--6592.

\bibitem{wang2022full}
Z.~Wang, B.~Zhao, P.~Zhang, L.~Yao, Q.~Wang, B.~Li, M.~Q.-H. Meng, and Y.~Hu, ``Full-coverage path planning and stable interaction control for automated robotic breast ultrasound scanning,'' \emph{IEEE Transactions on Industrial Electronics}, vol.~70, no.~7, pp. 7051--7061, 2022.

\bibitem{yang2021automatic}
C.~Yang, M.~Jiang, M.~Chen, M.~Fu, J.~Li, and Q.~Huang, ``Automatic 3-d imaging and measurement of human spines with a robotic ultrasound system,'' \emph{IEEE Transactions on Instrumentation and Measurement}, vol.~70, pp. 1--13, 2021.

\bibitem{hennersperger2016towards}
C.~Hennersperger, B.~Fuerst, S.~Virga, O.~Zettinig, B.~Frisch, T.~Neff, and N.~Navab, ``Towards mri-based autonomous robotic us acquisitions: a first feasibility study,'' \emph{IEEE transactions on medical imaging}, vol.~36, no.~2, pp. 538--548, 2016.

\bibitem{al2021autonomous}
L.~Al-Zogbi, V.~Singh, B.~Teixeira, A.~Ahuja, P.~S. Bagherzadeh, A.~Kapoor, H.~Saeidi, T.~Fleiter, and A.~Krieger, ``Autonomous robotic point-of-care ultrasound imaging for monitoring of covid-19--induced pulmonary diseases,'' \emph{Frontiers in Robotics and AI}, vol.~8, p. 645756, 2021.

\bibitem{shida2023automated}
Y.~Shida, S.~Kumagai, R.~Tsumura, and H.~Iwata, ``Automated image acquisition of parasternal long-axis view with robotic echocardiography,'' \emph{IEEE Robotics and Automation Letters}, 2023.

\bibitem{mnih2015human}
V.~Mnih, K.~Kavukcuoglu, D.~Silver, A.~A. Rusu, J.~Veness, M.~G. Bellemare, A.~Graves, M.~Riedmiller, A.~K. Fidjeland, G.~Ostrovski \emph{et~al.}, ``Human-level control through deep reinforcement learning,'' \emph{nature}, vol. 518, no. 7540, pp. 529--533, 2015.

\bibitem{yang2021searching}
X.~Yang, Y.~Huang, R.~Huang, H.~Dou, R.~Li, J.~Qian, X.~Huang, W.~Shi, C.~Chen, Y.~Zhang \emph{et~al.}, ``Searching collaborative agents for multi-plane localization in 3d ultrasound,'' \emph{Medical Image Analysis}, vol.~72, p. 102119, 2021.

\bibitem{alansary2019evaluating}
A.~Alansary, O.~Oktay, Y.~Li, L.~Le~Folgoc, B.~Hou, G.~Vaillant, K.~Kamnitsas, A.~Vlontzos, B.~Glocker, B.~Kainz \emph{et~al.}, ``Evaluating reinforcement learning agents for anatomical landmark detection,'' \emph{Medical image analysis}, vol.~53, pp. 156--164, 2019.

\bibitem{sahba2008application}
F.~Sahba, H.~R. Tizhoosh, and M.~M. Salama, ``Application of reinforcement learning for segmentation of transrectal ultrasound images,'' \emph{BMC medical imaging}, vol.~8, pp. 1--10, 2008.

\bibitem{liu2020ultrasound}
T.~Liu, Q.~Meng, A.~Vlontzos, J.~Tan, D.~Rueckert, and B.~Kainz, ``Ultrasound video summarization using deep reinforcement learning,'' in \emph{Medical Image Computing and Computer Assisted Intervention--MICCAI 2020: 23rd International Conference, Lima, Peru, October 4--8, 2020, Proceedings, Part III 23}.\hskip 1em plus 0.5em minus 0.4em\relax Springer, 2020, pp. 483--492.

\bibitem{ning2021autonomic}
G.~Ning, X.~Zhang, and H.~Liao, ``Autonomic robotic ultrasound imaging system based on reinforcement learning,'' \emph{IEEE Transactions on Biomedical Engineering}, vol.~68, no.~9, pp. 2787--2797, 2021.

\bibitem{li2023rl}
K.~Li, A.~Li, Y.~Xu, H.~Xiong, and M.~Q.-H. Meng, ``Rl-tee: Autonomous probe guidance for transesophageal echocardiography based on attention-augmented deep reinforcement learning,'' \emph{IEEE Trans. Autom. Sci. Eng.}, 2023.

\bibitem{mnih2013playing}
V.~Mnih, K.~Kavukcuoglu, D.~Silver, A.~Graves, I.~Antonoglou, D.~Wierstra, and M.~Riedmiller, ``Playing atari with deep reinforcement learning,'' \emph{arXiv preprint arXiv:1312.5602}, 2013.

\bibitem{van2016deep}
H.~Van~Hasselt, A.~Guez, and D.~Silver, ``Deep reinforcement learning with double q-learning,'' in \emph{Proceedings of the AAAI conference on artificial intelligence}, vol.~30, no.~1, 2016.

\bibitem{wang2016dueling}
Z.~Wang, T.~Schaul, M.~Hessel, H.~Hasselt, M.~Lanctot, and N.~Freitas, ``Dueling network architectures for deep reinforcement learning,'' in \emph{International conference on machine learning}.\hskip 1em plus 0.5em minus 0.4em\relax PMLR, 2016, pp. 1995--2003.

\bibitem{staal2007automatic}
J.~Staal, B.~van Ginneken, and M.~A. Viergever, ``Automatic rib segmentation and labeling in computed tomography scans using a general framework for detection, recognition and segmentation of objects in volumetric data,'' \emph{Medical image analysis}, vol.~11, no.~1, pp. 35--46, 2007.

\bibitem{salehi2015patient}
M.~Salehi, S.-A. Ahmadi, R.~Prevost, N.~Navab, and W.~Wein, ``Patient-specific 3d ultrasound simulation based on convolutional ray-tracing and appearance optimization,'' in \emph{MICCAI 2015: 18th International Conference, Munich, Germany, October 5-9, 2015, Proceedings, Part II 18}.\hskip 1em plus 0.5em minus 0.4em\relax Springer, 2015, pp. 510--518.

\bibitem{schaul2015prioritized}
T.~Schaul, J.~Quan, I.~Antonoglou, and D.~Silver, ``Prioritized experience replay,'' \emph{arXiv preprint arXiv:1511.05952}, 2015.

\end{thebibliography}

\end{document}